\documentclass[letterpaper]{article} 
\usepackage{aaai2026}  
\usepackage{times}  
\usepackage{helvet}  
\usepackage{helvet}  
\usepackage{amsmath}
\usepackage{amssymb}
\usepackage{courier}  
\usepackage[hyphens]{url}  
\usepackage{graphicx} 
\usepackage{natbib}  
\usepackage{caption} 
\usepackage{algorithm}
\usepackage{algorithmic}
\usepackage{amsmath}
\usepackage[table]{xcolor}
\usepackage{multirow}
\usepackage{appendix}
\nocopyright  
\usepackage{newfloat}
\usepackage{listings}
\DeclareCaptionStyle{ruled}{labelfont=normalfont,labelsep=colon,strut=off} 
\usepackage[svgnames]{xcolor} 
\usepackage[most]{tcolorbox}  

\newtcolorbox{notebox}[2][]{%
    enhanced,
    breakable,                  
    colback=white,              
    colframe=#2!80!black,       
    boxrule=0.5pt,              
    arc=3mm,                    
    outer arc=3mm,
    left=6mm, right=6mm, top=6mm, bottom=6mm, 
    boxsep=0pt,
    title={#1},                 
    fonttitle=\bfseries\sffamily,
    coltitle=white,
    boxed title style={
        colback=#2!80!black,
        boxrule=0pt,
        arc=2mm,
        outer arc=2mm
    },
    attach boxed title to top left={yshift=-4mm, xshift=5mm},
    drop shadow={#2!50!gray},   
}

\lstdefinestyle{mystyle}{
    basicstyle=\ttfamily\footnotesize,
    backgroundcolor=\color{black!3},
    frame=single,
    framesep=5pt,
    framerule=0.5pt,
    breaklines=true
}

\floatstyle{ruled}
\newfloat{listing}{tb}{lst}{}
\floatname{listing}{Listing}

\title{(P)rior(D)yna(F)low: A Priori Dynamic Workflow Construction via Multi-Agent Collaboration}
\author{
    Yi Lin,
    Lujin Zhao, 
    Yijie Shi,   
}
\affiliations{
    School of Cyberspace Security, Beijing University of Posts and Telecommunications \\
 \{yilin, zlujin, yijieshi2000\}@bupt.edu.cn
}

\usepackage{bibentry}

\begin{document}

\maketitle

%

\begin{abstract}
Recent studies have shown that carefully designed workflows coordinating large language models (LLMs) significantly enhance task-solving capabilities compared to using a single model. While an increasing number of works focus on autonomous workflow construction, most existing approaches rely solely on historical experience, leading to limitations in efficiency and adaptability. We argue that while historical experience is valuable, workflow construction should also flexibly respond to the unique characteristics of each task. To this end, we propose an a priori dynamic framework for automated workflow construction. Our framework first leverages Q-table learning to optimize the decision space, guiding agent decisions and enabling effective use of historical experience. At the same time, agents evaluate the current task progress and make a priori decisions regarding the next executing agent, allowing the system to proactively select the more suitable workflow structure for each given task. Additionally, we incorporate mechanisms such as cold-start initialization, early stopping, and pruning to further improve system efficiency. Experimental evaluations on four benchmark datasets demonstrate the feasibility and effectiveness of our approach. Compared to state-of-the-art baselines, our method achieves an average improvement of 4.05\%, while reducing workflow construction and inference costs to only 30.68\%–48.31\% of those required by existing methods.
\end{abstract}

\begin{links}
    \link{Code}{https://github.com/L1n111ya/PriorDynaFlow}
\end{links}

\section{Introduction}
Driven by technological advances, large language models (LLMs) have become powerful tools in reasoning, code generation, and mathematical problem-solving. Yet, as their capabilities are further explored, limitations—such as hallucinations and task-specific inefficiencies—have become increasingly evident. While LLMs excel at generating fluent text or solving isolated problems, they often struggle with compositional reasoning and cross-domain generalization. These challenges are particularly pronounced in complex, multi-step tasks requiring contextual awareness, where models may produce inconsistent outputs or fail to prioritize critical sub-problems.To address these issues, methods such as problem decomposition~\cite{MathAgent}, reflection~\cite{Reflextion}, and chain-of-thought prompting~\cite{CoT} have been proposed, showing promise in improving LLM performance. However, these approaches primarily mitigate surface-level errors rather than addressing fundamental architectural limitations—such as the inability to model dynamic task dependencies or adapt to novel problem structures.


\begin{figure}[t]
\centering
\includegraphics[width=0.9\columnwidth]{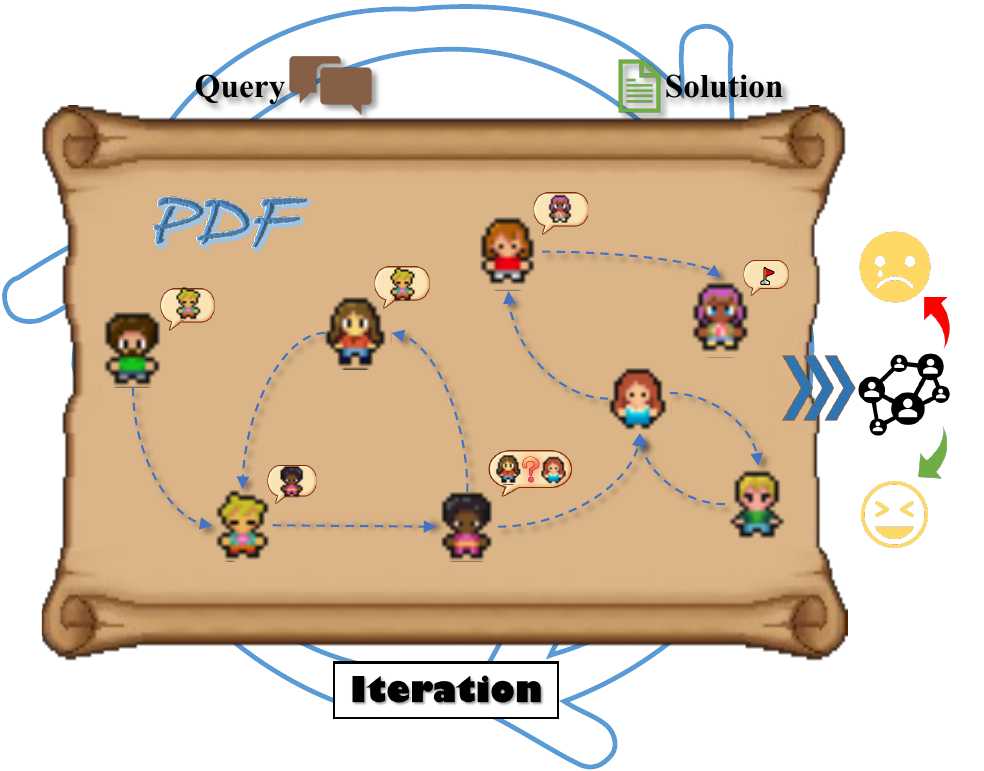} 
\caption{\textbf{Overview of PriorDynaFlow.} The framework dynamically operates the multi-agent system by enabling autonomous decision-making among agents, while Q-learning is employed to constrain and guide the decision space. For brevity, we denote it as PDF.}
\label{fig}
\end{figure}

In response, multi-agent systems have been increasingly recognized as a promising paradigm for addressing complex, real-world tasks. By decomposing problems into specialized sub-tasks and enabling collaborative decision-making, multi-agent frameworks can overcome LLMs' single-pass inference limitations. For example, \cite{ChatDev} demonstrated that assigning distinct roles to agents (e.g., planner, executor, critic) significantly improves performance on multi-hop reasoning tasks. Nevertheless, the widespread adoption of multi-agent systems faces two critical barriers: (1) \textbf{Design Complexity}: Constructing efficient agent workflows requires domain expertise and labor-intensive manual configuration, limiting scalability. (2) \textbf{Static Adaptability}: Most existing frameworks rely on predefined, rigid workflows that lack flexibility to adjust to problem-specific dynamics (e.g., varying task dependencies or resource constraints).

Recent efforts have focused on automating workflow construction to reduce design costs. For instance, \cite{AFlow} proposed a Monte Carlo Tree Search (MCTS)-based autonomous framework, while another work by \cite{AgentPrune} introduced a directed acyclic graph (DAG) pruning algorithm to optimize agent collaboration. \cite{DyLAN} conceptualized multi-agent systems as feedforward neural networks (FNNs), where an LLM Ranker selects layer-wise nodes to dynamically construct workflows. However, their workflow construction is typically a posteriori, and most approaches focus on static workflows. 

To address this challenge, we propose PDF, an a priori dynamic multi-Agent framework capable of autonomously constructing general-purpose workflows. Users define agent roles and behaviors, while PDF autonomously handles workflow construction and execution. Our approach emphasizes a priori workflow generation: after completing its action, an agent autonomously selects the next agent(s) or terminates the process, enabling proactive decisions based on task progress and learned experience.PDF employs Q-Learning with a reward mechanism to assign and update Q-values, constraining agent decisions to high-performing paths via the Q-table. To address the cold-start problem (initially empty Q-table), we maximize the decision space during early training, allowing any agent to be selected as the next node. This accelerates Q-table convergence. Mechanisms like early stopping and pruning further ensure efficiency and stability.

Our contributions are threefold:

\begin{itemize}
    \item \textbf{PDF Framework}: An a priori dynamic multi-Agent framework leveraging autonomous decision-making and Q-Learning to construct both general-purpose and task-specific workflows autonomously.
    \item \textbf{Theoretical Justification}: We formalize the reward function design in our Q-learning algorithm through mathematical proofs.
    \item \textbf{Empirical Validation}: Extensive evaluations on four benchmark datasets (spanning mathematics and code generation) demonstrate a 4.05\% average performance improvement over state-of-the-art baselines while reducing construction/inference costs to 30.68–48.31\% of existing solutions.
\end{itemize}

\section{Related Work}

Recently, the research on large-scale model Agents has become increasingly in-depth, and there have been many successful cases of applying it to various real-world fields(\cite{Code-Generation-with-AlphaCodium}, \cite{HAIChart}, \cite{Voyager}, \cite{Unleashing}, \cite{MobileExperts}, \cite{MetaGPT}, \cite{Lemur}, \cite{HAIChart}, \cite{Deeply}), such as code generation, mathematics, reasoning, and other domains.

\subsubsection{LLM-agent Workflow} \cite{SC} refers to the ability to generate multiple reasoning paths and aggregate results to output stable and logically closed-loop answers. \cite{Self-Refine} proposes an iterative refinement framework via self-feedback, leveraging a single LLM without extra data to enhance output quality. \cite{CodeAgent} enhances code generation capabilities through an agent system that integrates tool invocation, enabling it to tackle coding challenges at the real codebase level and achieve full-process automation from task planning to code implementation. \cite{ChatDev} constructing a multi-agent system with communication and collaboration capabilities, which realizes full-process automated software development from requirement analysis to code implementation by simulating the division of labor and interaction patterns of human software development teams. \cite{Debugger-LLM} introduced a human-inspired workflow in which code is debugged incrementally, enabling large language models to generate more accurate and reliable code. 

\subsubsection{Mutil-Agent Reinforcement Learning} \cite{AutoFlow} AutoFlow proposes a natural-language workflow generation framework with iterative optimization. \cite{AlphaGo} and \cite{Dota2} applied deep neural networks with MCTS and multi-agent RL to defeat human professionals in Go and Dota 2. \cite{Reflextion} proposed the Reflexion framework, which enhances the continuous learning and adaptive capabilities of language agents in complex tasks by enabling them to conduct self-reflection in natural language and optimize decision-making processes using reinforcement learning. \cite{AFlow} proposed an automated workflow construction method based on Monte Carlo Tree Search (MCTS). \cite{AgentPrune} argued that agent workflows can be modeled as topological structures, and presented a novel pruning framework leveraging masking techniques to dynamically generate directed acyclic graphs (DAGs) in an automated manner. \cite{DyLAN} introduced a neural network-inspired dynamic workflow paradigm, where agent collaboration is modeled as a feedforward neural network (FNN), and node transitions are determined through agent importance scoring.

\begin{figure*}[t]
\centering
\includegraphics[width=0.8\textwidth]{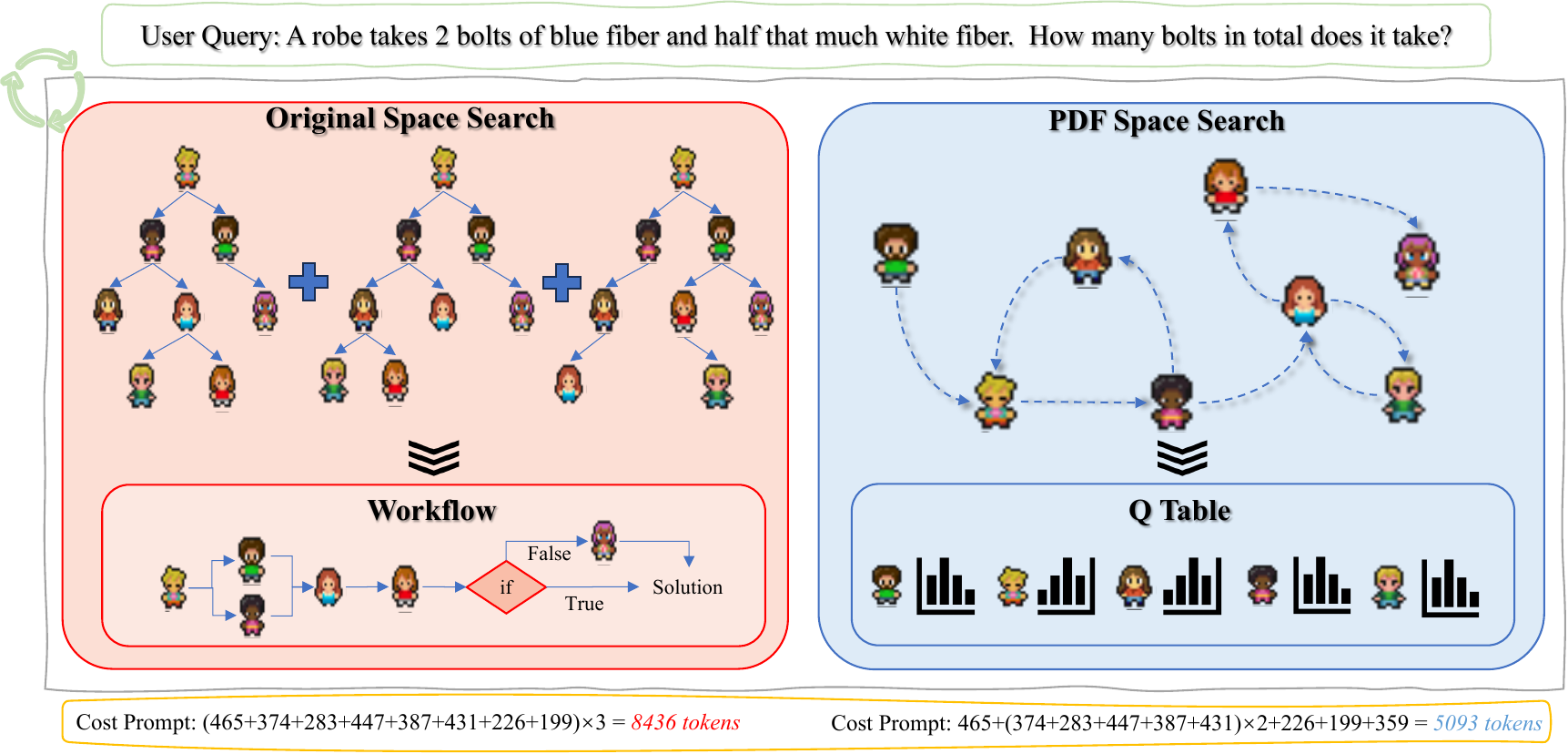} 
\caption{\textbf{The overall of our PDF framework.} The previous spatial search methods were usually traded for resources and time, optimizing workflows through continuous experimentation, which resulted in erroneous attempts wasting resources. PDF obtains directed edges through agent autonomous decision-making, constructs a workflow graph, and optimizes the decision space through a reward mechanism, thereby saving resources and time consumption caused by many erroneous attempts.}
\label{figure2}
\end{figure*}

\section{Preliminary}

\subsection{Problem Formulation}

At its core, a dynamic workflow can be represented as a collection of nodes and directed edges. In contrast to traditional workflows, where edges are typically unweighted directed edges, dynamic workflows generally employ weighted directed edges, allowing for richer modeling of information flow and decision-making.

Below, we define several key components used throughout this paper:
\begin{itemize}
    \item \textbf{Node} \textit{N}: A node corresponds to a role in the workflow and is defined as a combination of a Large Language Model (LLM) and its designated action.
    \item \textbf{Prompt} \textit{P}: Prompts are divided into two types—system prompts, which provide general task instructions, and role prompts, which specify the behavior and responsibilities of a particular role.
    \item \textbf{Weighted Directed Edge} \textit{E}: A directed edge with an associated weight, indicating the direction of information flow from one role to another. The meaning of the weight varies depending on the system design.
\end{itemize}

In DyLAN, the weight represents an importance score, reflecting the relevance or necessity of the next role in completing the task. This directly influences the selection of the subsequent agent.In our method, the weight encodes the reward or penalty assigned to the transition from Role A to Role B, guiding the learning process through reinforcement signals.

The essence of a workflow as a structure defined by a set of nodes \textit{N} (representing agents) and edges \textit{E} (representing collaboration paths). In PDF, the autonomous decision-making space corresponds to the set of nodes \textit{N}, while agents dynamically generate the edge list \textit{E} based on their execution status and contextual requirements. Mathematically, the operational process of PDF can be represented as follows:
\begin{equation}
    W^* = \arg \max_{W\in S}(W, E, N)
\end{equation}

\subsection{Q-Learning}
\label{sub:qlearning}
Q-Learning, a temporal difference algorithm, estimates action values through agent-environment interactions. The Q-table is a 2D list in the form $Q[state][action]=value$, recording the quality of actions for each state. Specifically, we use the temporal difference learning method to update and estimate the Q-table:
\begin{equation}
\begin{split}
    Q_{new}(s_t, a_t) = & (1 - \alpha)Q(s_t, a_t) + \alpha [Reward_{s_t} + \\
            & \gamma \max_{a}Q(s_{t+1},a)]
\end{split}
\label{eq:qlearning}
\end{equation}
where $\alpha$ is the step size used to control the weight of each edge's impact on the Q-table; $\gamma$ is a discount factor, and the existence of a discount factor affects the updating of the Q-table by future rewards, thereby enabling the Q-table to learn long-term dependencies. In a deterministic environment, Q-Learning can converge to the optimal strategy, and the optimization objective is:
\begin{equation}
    \pi^*(s) = \arg\max Q^*(s,a),s\in S
\end{equation}
The specific proof of the convergence of the Q-Learning algorithm is provided in appendix.

\subsection{$\epsilon$-Greedy Strategy}Due to the limitations of Q-Learning itself, the decision space of the agent is the node space with higher known value estimates, but the known value estimates may not be accurate. Strictly following this strategy will make the algorithm a greedy algorithm, which is prone to getting stuck in local optima. In order to ensure that a larger space can be searched, we introduce the $\epsilon$-greedy strategy when obtaining the decision space. Specifically, when obtaining the decision space of the current agent, an action will be randomly selected with an epsilon iteration probability, and together with the $topK$ actions with the highest Q value, the decision space will be formed, so that the agent will reserve a certain probability to explore new actions and discover better decision spaces.

\subsection{Roles}
\label{sub:roles}
We strongly align with \cite{ChatDev} approach, emphasizing the critical importance of role definitions in multi-agent workflows. Properly designed roles—accurate, rational, and semantically rich—substantially enhance problem-solving efficacy. 

By explicitly defining roles, our framework requires users to specify only the behaviors and responsibilities of each role, eliminating the need for manual design of interaction protocols or workflow architectures. This paradigm significantly simplifies and accelerates workflow construction. In this work, we define nine specialized roles: (1) Algorithm Designer,
(2) Researcher, (3) Programming Expert, (4)Code Reviewer, (5)Test Engineer, (6) Mathematician, (7)Programming Assistant, (8) Data Analyst, (9)Inspector. Each role operates autonomously, executing tasks aligned with its designated expertise (e.g., code generation by the Programming Expert, validation by the Validator). Collaborative problem-solving emerges through role-specific interactions, enabling holistic task completion. 

\section{A Priori Nature}
Previous methods(\cite{AFlow}, \cite{DyLAN}) for autonomous workflow construction are fundamentally a posteriori, relying entirely on historical "experience". The experience gained in the current iteration cannot influence decisions made within that same iteration. Instead, these methods continuously explore the workflow space to discover better configurations—effectively accumulating more experience over time. However, such experience-driven approaches may be ineffective when prior knowledge is not applicable to the current task.

In autonomous workflow construction, the key question should not be what the next node of the current agent is, but rather, what the correct next node is given the current task state. Our method is built upon this principle, making decisions based on the current state of the task in an a priori manner. The decision space of each agent is optimized using a Q-table, which is inherently a posteriori—it represents the experience accumulated through prior exploration. However, the decision-making process itself is a priori, meaning it is made proactively based on real-time context rather than retrospective evaluation.

Specifically, each agent is assigned a decision space (i.e., a set of possible next agents to choose from, derived from past experience) and a given task. The agent first performs its designated role to complete the corresponding part of the task. Upon completion, it evaluates the current progress toward task resolution and receives information about the roles of its available collaborators (i.e., what actions they can perform). Based on this information, the agent then selects the most appropriate next node (or multiple nodes, if configured to do so).

This autonomous decision-making process is a priori in nature—unlike existing methods, our framework allows decisions to be made within the current iteration without requiring trial-and-error or post-hoc evaluation. As a result, the agent’s decision can directly influence the ongoing workflow execution, enabling more adaptive and efficient task completion.

\section{PriorDynaFlow}

The central philosophy of our framework is agent autonomous decision-making. Agents should generate edge sequences based on their runtime status while fulfilling their core responsibilities, thereby achieving dynamic workflows. Meanwhile, we incorporate Q-Learning to constrain the autonomous decision-making space of agents during framework execution. By restricting this space, workflow variability is confined to high-quality subspaces, ensuring smooth multi-agent collaboration. As depicted in Figure \ref{figure2}, our method demonstrates remarkable efficiency in exploring the search space when autonomously constructing multi-agent workflows, outperforming existing methods. For each user query, only a single exploration is required to facilitate iterations, eliminating the necessity for multiple exploratory and comparative iterations.

Before executing its task, each agent selects the top-k nodes (potential successor agents) from the current state and Q-table as its autonomous decision-making space. After completing its behavior, the agent dynamically chooses the next agent to collaborate with, forming an edge between them. Certain agents are equipped with termination capabilities, allowing them to decide whether to conclude the workflow. If no agent terminates the workflow or unexecuted edges remain, the process repeats iteratively. In a user query process, we usually only need to go through the above process once to obtain a set of edges and a list of edge weights. Based on this list of edge weights, we can update the Q-table.

\subsection{Decision Space Estimate}
As shown in Algorithm 1 in the Appendix, our framework employs Q-Learning to estimate and refine agent decision spaces throughout the workflow construction process. Each agent’s decision space is derived from a Q-table, which encodes the expected utility of selecting a particular action (i.e., choosing a specific next agent) under a given task state. Over time, this Q-table is updated based on task execution feedback, enabling agents to make increasingly informed and adaptive decisions.

However, at the initial stage of system execution, the Q-table is typically initialized randomly for all state-action pairs. If agents were to select actions based solely on these untrained Q-values, the resulting decisions would likely be suboptimal or even harmful to the learning process. To address this challenge and inspired by \cite{DeepSeek-R1}, we introduce a cold-start mechanism into PDF.

\begin{figure*}[t]
\centering
\includegraphics[width=0.9\textwidth]{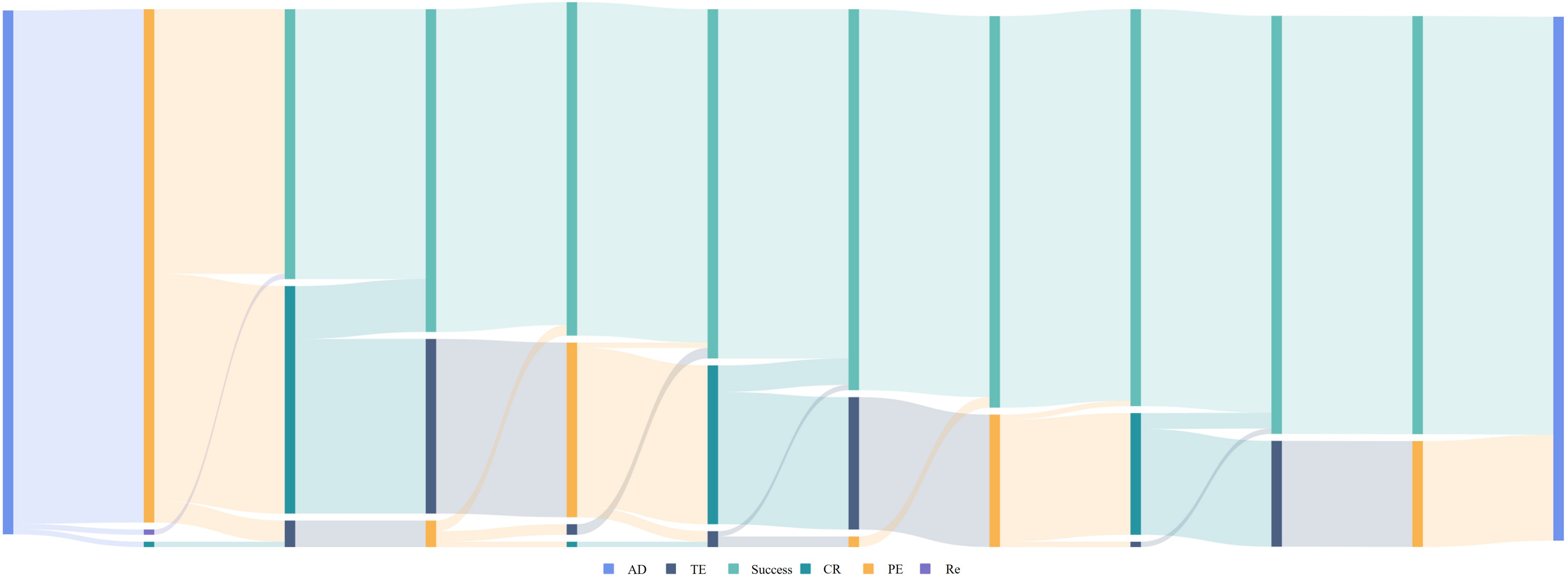} 
\caption{\textbf{A Sankey diagram showing the nodes traversed in each round of tasks.} Among the 161 tasks in the HumanEval task, this diagram illustrates the dynamically constructed workflows by our method for different tasks. Here, AD stands for Algorithm Designer, PE for Programming Expert, CR for Code Reviewer, TE for Test Engineer, Re for Researcher(mole), and Success indicates the end of the workflow. The last column represents the end of the workflow.}
\label{sankey}
\end{figure*}

During the cold-start phase, we temporarily expand each agent’s decision space to include the full set of available actions $\Omega$, rather than restricting it based on early, unreliable Q-values. Although this may introduce some instability in the early stages of workflow execution, it ensures that agents can explore a wide range of possible actions and avoid premature convergence to suboptimal paths. As the Q-table accumulates meaningful reward signals, the decision space gradually narrows to include only the most promising actions.

Specifically, we design a reward mechanism tailored for autonomous multi-Agent workflow construction. This mechanism assigns a reward weight to each directed edge whenever an agent is invoked. The reward is determined based on the agent’s action and its corresponding performance in completing the assigned task. In this formulation, the two endpoints of an edge represent the current state and the selected action, respectively. Using these reward signals, the Q-table is updated via the temporal difference learning approach as shown in Equation~\ref{eq:qlearning}.

To further refine the decision-making process, we constrain each agent’s choice to the top-k actions (i.e., next agents) with the highest Q-values under the current state.

\begin{equation}
\begin{split}
    \mathcal{A}_{top-k}(s)=\arg\max_{a\in \mathcal{A}(s)}^{(k)}Q(s,a)
\end{split}
\end{equation}
where $\mathcal{A}(s)$ represents the complete decision space. This ensures that the agent operates within a high-quality, compact decision space while still maintaining flexibility and adaptability.

\subsection{Reward Mechanism Design} Firstly, we believe that a good Agent workflow should be as short as possible and use as few nodes as possible. Therefore, in order to ensure that the optimization objectives of Q-Learning can meet the above requirements, our designed reward mechanism introduces execution penalty and edge penalty; Secondly, as discussed in section~\ref{sub:roles}, we believe that different agents should play different roles in multi-agent workflows. Therefore, in order to better solve the problem, theoretically, the starting nodes of directed edges should be different, because the behavior performed by the same node is the same. Completing the same behavior twice is repetitive and will only consume computing resources; In addition, we believe that the execution penalties for different nodes should be different, and nodes with higher efficiency in solving user problems should receive fewer execution penalties. We should encourage the Agent workflow to use more efficient nodes, so we have introduced success rate based rewards; Finally, after completing user tasks, a large amount of rewards should be obtained in order to make Q-Learning's spatial estimation tend to solve user tasks. In summary, the design of the reward mechanism mainly includes five aspects:
\begin{itemize}
    \item Punishment for the execution node itself
    \item Punishment for a repeating node
    \item Punishment for the number of edges
    \item The rewards brought by the success rate of this state
    \item The large rewards brought by completing user tasks
\end{itemize}

The core algorithm of the reward mechanism is shown in Algorithm 2 in the Appendix.

\rowcolors{2}{gray!25}{} 
\begin{table*}[h!t]
    \centering
    \setlength{\tabcolsep}{10pt} 
    \begin{tabular}{l | c c c c | c}
        \hline\hline
        \rowcolor{lightgray} 
        \textbf{Method} & 
        \textbf{HumanEval} & 
        \textbf{MBPP} & 
        \textbf{GSM8K} & 
        \textbf{Math} &
        \textbf{Avg.}\\ 
        \hline
        Original & 
        $73.2$ & 
        $80.6$ & 
        $94.5$ & 
        $68.5$ &
        $79.2$\\

        \hline
        
        CoT & 
        $90.06$ \, {{\footnotesize \itshape $ \uparrow 16.86 $}} & 
        $80.80$ \, {{\footnotesize \itshape $ \uparrow 0.20 $}} & 
        $93.89$ \, { \footnotesize \itshape $ \downarrow 0.61 $} &
        $78.67$ \, {{\footnotesize \itshape $ \uparrow 10.17 $}} & 
        $85.86$ \\

        CoT SC & 
        $90.68$ \, {{\footnotesize \itshape $ \uparrow 17.48 $}} & 
        $81.33$ \, {{\footnotesize \itshape $ \uparrow 0.73 $}} & 
        $93.97$ \, { \footnotesize \itshape $ \downarrow 0.53 $} &
        $84.13$ \, {{\footnotesize \itshape $ \uparrow 15.63 $}} & 
        $87.53$ \\

        Reflextion & 
        $93.16$ \, {{\footnotesize \itshape $ \uparrow 19.96 $}} & 
        $83.98$ \, {{\footnotesize \itshape $ \uparrow 3.38 $}} & 
        $93.55$ \, { \footnotesize \itshape $ \downarrow 0.95 $} & 
        $80.00$ \, {{\footnotesize \itshape $ \uparrow 11.50 $}} &
        $87.67$\\

        \hline
        
        AgentPrune & 
        $91.87$ \, {{\footnotesize \itshape $ \uparrow 18.67 $}} & 
        $82.20$ \, { \footnotesize \itshape $ \uparrow 1.60 $}& 
        \textbf{97.18} \, { \footnotesize \itshape $ \uparrow 2.68 $}& 
        $80.60$ \, { \footnotesize \itshape $ \uparrow 12.10 $} &
        $87.96$\\
        
        AFlow & 
        $88.20$ \, { \footnotesize \itshape $ \uparrow 15.00 $} & 
        $78.20$ \, { \footnotesize \itshape $ \downarrow 2.40 $} & 
        $92.88$ \, { \footnotesize \itshape $ \downarrow 1.62 $} & 
        $80.33$ \, { \footnotesize \itshape $ \uparrow 11.83 $} & 
        $84.91$ \\

        DyLAN & 
        $93.73$ \, { \footnotesize \itshape $ \uparrow 20.53 $} & 
        $84.60$ \, { \footnotesize \itshape $ \uparrow 4.00 $} & 
        $91.90$ \, { \footnotesize \itshape $ \downarrow 2.60 $} & 
        $84.33$ \, { \footnotesize \itshape $ \uparrow 15.83 $} &  
        $88.64$ \\

        \hline

        \textbf{Ours w/o P}  & 
        $87.58$ \, { \footnotesize \itshape $ \uparrow 14.38 $} & 
        $80.60$ \, { \footnotesize \itshape $ \uparrow 0.00 $} & 
        $95.49$ \, { \footnotesize \itshape $ \uparrow 0.99 $} & 
        $86.50$ \, { \footnotesize \itshape $ \uparrow 18.00 $} &
        $87.54$\\
        
        \textbf{Ours[Full]}  & 
        \textbf{93.90} \, { \footnotesize \itshape $ \uparrow 20.70 $} & 
        \textbf{89.40} \, { \footnotesize \itshape $ \uparrow 8.80 $} & 
        $96.36$ \, { \footnotesize \itshape $ \uparrow 1.96 $} & 
        \textbf{89.10} \, { \footnotesize \itshape $ \uparrow 20.60 $} &
        \textbf{92.19}\\
        \hline\hline
    \end{tabular}
    \caption{Performance. We use Qwen2.5-Max as the base model for all experiments. To verify whether our method is model-agnostic, we provide a multi-model comparison in the appendix to demonstrate the generalizability of our approach across different architectures. An ablation study is further conducted to evaluate the contribution of the "Prior" component, where we remove this key module and analyze its impact on performance.}
    \label{tab:performance}
\end{table*}

\subsection{Pruning} 
Due to the inherent limitations of large language models (LLMs) and the quality or completeness of user-provided information, certain user queries cannot be effectively addressed by the current workflow space. In such cases, the dynamic workflow may fall into a loop where new agents (and corresponding edges) are continuously added without reaching termination—a common issue in adaptive multi-agent systems. Since the reward mechanism penalizes path length, excessively long workflows incur substantial cumulative penalties, which can significantly distort the estimation of the decision space and hinder the learning process.

To mitigate this issue, we introduce a pruning mechanism. Specifically, when the workflow enters a state of continuous penalization—indicated by the total accumulated reward falling below a predefined threshold—we enforce an early termination of the workflow. Furthermore, the entire path generated during this attempt is discarded (i.e., "pruned"), and the corresponding list of directed, weighted edges is not used to update the Q-table. This ensures that poorly performing or non-convergent workflows do not negatively impact the learning of future decision-making policies.

\section{Experiments}
In this section, we formulate the following research questions and validate our responses through rigorous experiments: (\textbf{RQ1}) How does PDF perform in terms of task-solving performance and computational efficiency when addressing user queries? (\textbf{RQ2}) Does PDF genuinely estimate a more effective decision space? (\textbf{RQ3}) Why Does A Priori Adaptability Improve Dynamic Workflow Construction? Can our method generalize across diverse task types through dynamic workflow construction?(\textbf{RQ4})

\subsection{Experiments Setup}
\subsubsection{Datasets} 
To validate the effectiveness of our approach, we conduct experiments on four benchmark datasets covering two domains: code generation and mathematical reasoning. Specifically, for code generation, we evaluate on HumanEval\cite{HumanEval} and MBPP\cite{MBPP}, two widely used programming benchmarks, and report results based on their test splits. For mathematical reasoning, we use the GSM8K\cite{GSM8K} and MATH\cite{Math} datasets in their entirety to assess performance across a range of problem complexities. These datasets provide a comprehensive evaluation of PDF’s capability in solving diverse user tasks under different application scenarios.

\subsubsection{Baselines}
We compare our approach against a set of baseline methods, which include both manually designed and automatically generated workflow-based strategies. Specifically, for the category of manually constructed workflows, we adopt Reflection\cite{Reflextion} and Chain-of-Thought\cite{CoT} as representative baselines. For the category of automatically generated workflows, we select AgentPrune\cite{AgentPrune}, DyLAN\cite{DyLAN}, and AFlow\cite{AFlow} as state-of-the-art methods to benchmark against. Covering diverse paradigms, these baselines provide a robust benchmark for evaluating PDF.
\begin{figure}[t]
\centering
\includegraphics[width=0.9\columnwidth]{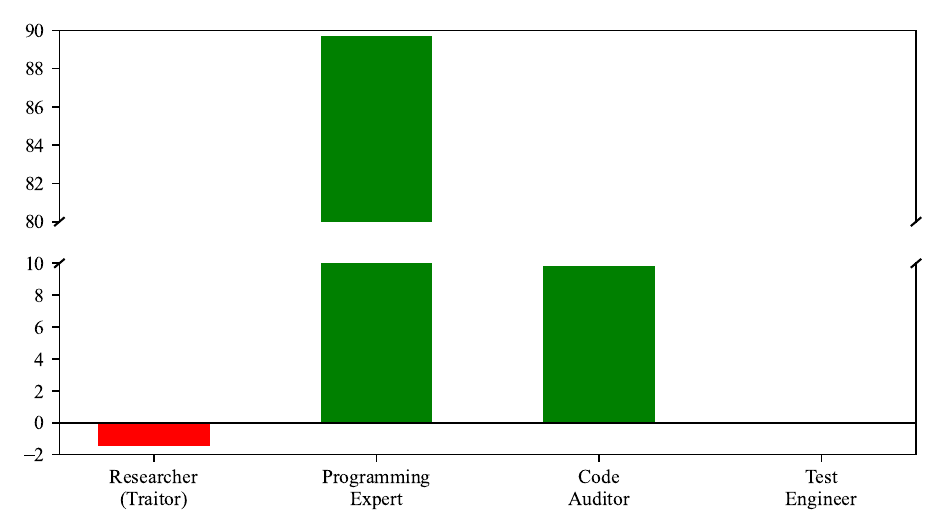} 
\caption{ Estimated action space values for the "Algorithm Designer" state after 161 iterations on the HumanEval dataset, with the first 30 iterations designated as cold start.}
\label{fig4}
\end{figure}
\subsubsection{Metrics}
For evaluation against the four baseline methods mentioned above, we adopt the pass@1\cite{pass@1} as our primary performance measure. This metric evaluates whether the system successfully solves the given task on its first attempt, without any retries. We apply this metric consistently across both code generation and mathematical reasoning tasks, granting each workflow only one opportunity to resolve the problem. Notably, the cold-start phase is also included in this single attempt.

\subsection{Performance and Efficiency Evaluation(RQ1)}
We evaluate our method in terms of both task performance and computational efficiency, comparing it with two state-of-the-art approaches: DyLAN~\cite{DyLAN}, which models workflows as feedforward neural networks, and AgentPrune~\cite{AgentPrune}, which employs a pruning-based communication strategy. While these methods offer structural or efficiency benefits, they either incur high computational costs or lack dynamic and a priori decision-making capabilities.

Our method overcomes these limitations by integrating dynamic adaptability with a priori reasoning, enabling effective workflow construction without exhaustive exploration. As shown in Table~\ref{tab:performance}, our approach consistently outperforms all baselines across four benchmark datasets, achieving an average improvement of 4.05\% over the best-performing existing method. Notably, on challenging tasks such as code generation and mathematical reasoning, where base models struggle, our method achieves up to a 20\% performance gain, with significantly lower variance, indicating strong robustness and generalization.

In terms of efficiency, as shown in Table 1 in Appendix, our method reduces computational cost by 14.57–48.31\% compared to the most efficient baseline and by 51.79–69.32\% compared to the current state-of-the-art dynamic method. This demonstrates that our approach not only improves performance but also achieves substantial efficiency gains, striking a favorable balance between accuracy and resource consumption.

In summary, this analysis demonstrates that our method not only delivers superior performance but also achieves significant efficiency gains, making it a more practical and scalable solution for autonomous workflow construction.

\subsection{Decision Space Estimation Capability(RQ2)}
Dynamic workflow construction involves selecting the most suitable agent for the next step—essentially refining the decision space in real time. To evaluate whether our method can effectively distinguish between helpful and unhelpful agents, we introduce a “traitor role” into the experimental setup. This role appears normal but behaves adversarially. In the code generation task, we add a Researcher role alongside standard agents such as Algorithm Designer, Programming Expert, Code Reviewer, and Test Engineer. While the Researcher presents itself as capable of solving complex problems, it contributes little and may even hinder task completion.

As shown in Figure~\ref{fig4}, the estimated value of the Researcher remains consistently negative throughout training, reaching –1.484 by the end. In contrast, all legitimate roles maintain positive value estimates, typically above 5. Note that all roles start with an initial value of 0; negative values indicate low utility and reduced likelihood of being selected (see Supplementary Material for details). This result confirms that our framework can accurately assess the long-term utility of each agent based on historical performance—even identifying adversarial or ineffective ones.

\subsection{Impact of priori adaptability(RQ3)}
We argue that a priori decision-making is central to effective autonomous workflow construction, enabling the system to make timely, informed decisions by leveraging valuable historical experience. In contrast, most existing methods rely entirely on past experience, which is often accumulated through extensive trial and error. This not only leads to inefficiency during execution but also results in suboptimal workflow structures when applied to tasks that differ significantly from those in the training data.

To further validate the necessity of a priori adaptability, we conduct an ablation study by removing the a priori component from our framework, reducing it to a purely a posteriori approach. As shown in Figure~\ref{tab:performance}, performance drops on average by 4.65\% across four datasets, confirming the critical role of a priori decision-making in dynamic workflow construction.

Additionally, we evaluate our method on the HumanEval dataset, where task complexity varies significantly. As illustrated in Figure~\ref{sankey}, our framework dynamically adjusts the length of generated workflows, ranging from 2 to 11 steps, depending on the difficulty of the input task. This adaptability ensures that simpler tasks are completed efficiently, without unnecessary resource consumption, while more complex tasks are allocated sufficient steps to achieve a high-quality solution.

Crucially, our method leverages the Q-table to make a priori decisions about the most suitable workflow structure at each step. This allows the system to proactively avoid inefficient or inappropriate execution paths, significantly reducing the need for costly backtracking or repeated agent invocation. It represents a fundamental shift from reactive exploration to proactive planning, a key differentiator of our approach.


\subsection{Generalization Across Task Types(RQ4)}
\begin{figure}[t]
\centering
\includegraphics[width=0.85\columnwidth]{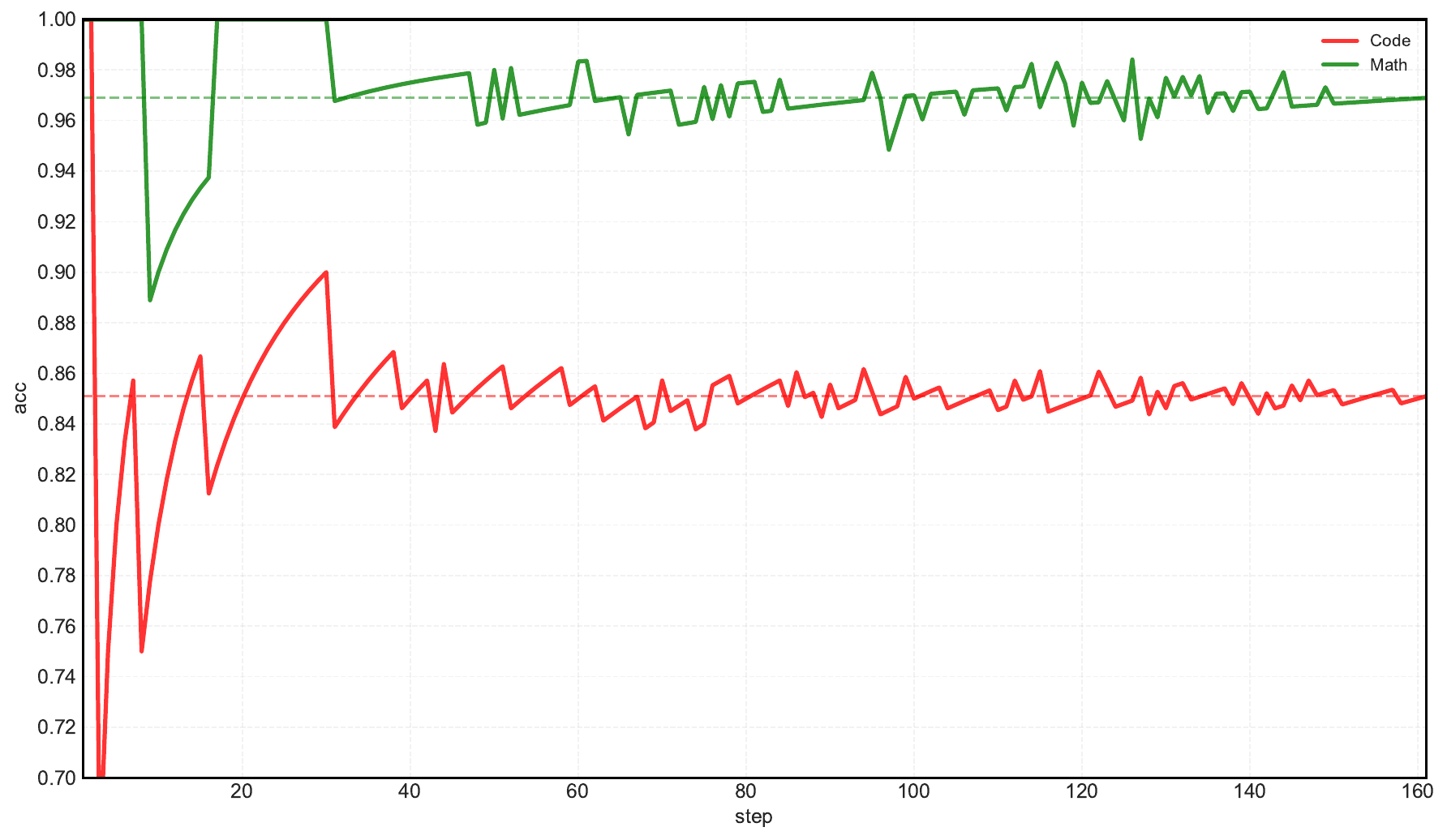} 
\caption{We combined the entire HumanEval dataset with 161 randomly selected tasks from Gsm8k to construct a mixed-task dataset, where coding tasks and mathematical tasks are interleaved.}
\label{fig5}
\end{figure}

Our method demonstrates strong adaptability by enabling generic dynamic workflow construction to address diverse tasks. To evaluate this capability, we define a unified framework incorporating six specialized roles: Planner, Programming Expert, Data Analyst, Inspector, Code Reviewer, and Test Engineer, which span generalist, math-specific, and code-specific functionalities. Experiments on mixed code-math tasks (Figure~\ref{fig5}) show that our framework achieves significant performance gains under a single generic workflow: a 11.89\% improvement in code tasks and a 2.39\% improvement in math tasks, comparable to the performance of dedicated workflow designs. In contrast, existing posteriori methods rely on fixed node selection based on historical experience, limiting their ability to dynamically adapt to task-specific requirements.

\section{Conclusion}
This paper introduces PDF, a prior, high-performance, and resource-efficient framework for autonomously constructing dynamic workflows by modeling them as role-edge sets, with the capability to build general-purpose workflows. Leveraging Q-Learning to estimate role-specific decision spaces, PDF enables agents to dynamically select optimal workflow edges. We detail the reward mechanism and Q-Learning-based estimation method. Experiments on four benchmarks demonstrate superior cross-domain performance, confirming PDF's efficacy in constructing high-quality workflows with minimal computational overhead.

\bigskip
\noindent

\appendix
\section*{\centering \Large \bfseries Appendix}  
\vspace{5pt}
\section{Additional Experimental Results}

\subsection{Hardware Specifications and Hyperparameters}
\subsubsection{Hardware Specifications}
In terms of hardware specifications, the experiments were conducted on a MacBook Pro (M1 Pro) equipped with 32GB of memory, a 512GB SSD, and running macOS 15.5 (Sequoia).

\subsubsection{Hyperparameters}
All experiments were conducted with a temperature of 0.1 and a max\_token of 2048.

\subsection{Baseline Methods}

This section introduces the baseline methods used in the experiments presented in the main text.

\subsubsection{CoT} Before the emergence of specialized reasoning models, large language models (LLMs) often struggled to provide accurate answers to complex problems. This limitation stems not only from the inherent constraints in their reasoning capabilities but also from the multi-step nature of such tasks, which demands structured and sequential thinking. To address this challenge, Chain-of-Thought (CoT)\cite{CoT} prompting has been introduced as a powerful technique that leverages in-context learning to guide LLMs toward more effective reasoning. In CoT prompting, models are instructed or prompted to explicitly generate intermediate reasoning steps before producing the final answer. This approach significantly improves the model’s ability to handle arithmetic, commonsense, and logical reasoning tasks. By encouraging the model to simulate a step-by-step thought process, CoT enhances both the accuracy and interpretability of the model's outputs. Furthermore, the generated intermediate steps offer users insights into how the model arrives at its conclusions, thereby increasing transparency and trust in the model’s decision-making process. Today, chain-of-thought reasoning has become a widely adopted strategy for enabling LLMs to tackle complex and multi-step tasks effectively.

\rowcolors{1}{gray!25}{}
\begin{table*}[htbp]
    \centering
    \setlength{\tabcolsep}{12pt} 
    \begin{tabular}{l|l|r|r|r}
        \hline
        \textbf{Dataset} & \textbf{Method} & \textbf{Prompt token} & \textbf{Completion token} & \textbf{Cost(\$)} \\
        \hline
        HumanEval & DyLAN & 1,817,938 & 740,794 & 11.475 \\
        \hline
        & AgentPrune & 1,192,938 & 396,732 & 6.672 \\
        \hline
         & Our        & 928,053   & 386,187 & 5.943 \\
        \hline
         GSM8K     & DyLAN & 4,980,841 & 3,244,003 & 43.096 \\
        \hline
        & AgentPrune & 3,523,234 & 723,942 & 15.376 \\
        \hline
             & Our        & 2,191,169 & 829,519 & 13.222 \\
        \hline
    \end{tabular}
    \caption{Comparison of token usage and cost between methods on different datasets.}
    \label{tab:token_cost}
\end{table*}

\begin{algorithm}[tb]
\caption{Decision Space Estimate Algorithm}
\label{Main Algorithm}
\begin{algorithmic}[1] 
\STATE Initialize Q-table $ Q(s, a) \gets 0 $ 
\STATE Initialize hyperparameters: learning rate $ \alpha $, discount factor $ \gamma $, exploration rate $ \epsilon $
\FOR{each training episode}
\STATE Initialize current state $ s \gets s_0 $, executed nodes $ E \gets \emptyset $, $ done \gets \text{False} $
\WHILE{not $ done $}
\STATE Get available actions: $ A(s) \gets \texttt{get\_nodes}(s) $
\IF{$ E $ is empty}
\STATE Remove END from $ A(s) $
\ENDIF
\STATE Select action $ a $ using $ \epsilon $-greedy:
\STATE $ a \gets \max_{a'} Q(s, a') + \text{random}(A(s))$
\STATE Execute node $ a $, observe reward $ r $ and adge $ e $
\STATE Update $ E \gets E \cup \{e\} $
\STATE Calculate Reward r: $r \gets reward(s, a, E)$
\STATE Update Q-value:
   $$
   Q(s, a) \gets (1 - \alpha)Q(s, a) + \alpha \left[ r + \gamma \max_{a'} Q(s', a') \right]
   $$
\STATE $ s \gets s' $
\ENDWHILE
\STATE Decay $ \epsilon $
\ENDFOR
\end{algorithmic}
\end{algorithm}

\subsubsection{Reflextion} The Reflexion\cite{Reflextion} framework introduces the novel concept of self-reflection, enabling agents to iteratively improve their performance through continuous interaction among three key roles: \textbf{Actor}, \textbf{Evaluator}, and \textbf{Self-Reflection}. In this paradigm, the Actor generates actions to perform the given task, the Evaluator assesses the outcomes of these actions, and the Self-Reflection component synthesizes insights from both the action and evaluation results. These insights are then stored in a long-term memory module, allowing the system to accumulate experience over time. This iterative process continues until the Actor’s actions successfully achieve the desired goal. By integrating self-awareness and memory into the decision-making loop, Reflexion enables agents to dynamically adapt and refine their strategies, thereby significantly enhancing overall task-solving capabilities.

\subsubsection{AtengPrune} AgentPrune\cite{AgentPrune} is a communication-efficient framework for multi-agent systems that introduces the concept of communication redundancy in the inter-agent interaction topology. It challenges the conventional assumption that transmitting large volumes of information across agents necessarily improves overall system performance. Instead, AgentPrune argues that significant redundancy exists in multi-agent communication, and eliminating this redundancy can lead to both improved efficiency and enhanced task-solving capabilities. The method models multi-agent communication as a spatio-temporal graph, where temporal edges represent sequential interactions over time, and spatial edges denote concurrent agent-to-agent communications. By applying both \textbf{temporal pruning} and \textbf{spatial pruning}, AgentPrune removes unnecessary communication pathways while preserving critical decision-making signals. This results in reduced communication overhead and more focused, effective collaboration among agents.

\subsubsection{AFlow} AFlow\cite{AFlow} introduces Monte Carlo Tree Search (MCTS) to systematically explore and identify optimal workflows. MCTS is a powerful heuristic search method that effectively balances exploration and exploitation within large or complex search spaces. By modeling the workflow construction process as a tree, AFlow enables intelligent backtracking and reuse of previously acquired knowledge. The tree structure allows nodes (i.e., agent actions or intermediate steps) to retain information from prior attempts—particularly failure experiences—when they are revisited. This mechanism helps the system avoid repeating past mistakes and guides the search toward more promising paths. As a result, AFlow achieves more stable and efficient workflow discovery compared to methods that lack memory-augmented exploration strategies.

\subsubsection{DyLAN} DyLAN\cite{DyLAN} is a dynamic multi-agent collaboration framework designed to organize and manage multiple large language models (LLMs) in a coordinated manner, enabling them to collaboratively solve complex tasks. The core mechanism of DyLAN lies in its importance scoring system, which evaluates the contribution of each agent’s output after every execution round. Based on these importance scores, DyLAN updates an LLM-based Ranker in a feedback loop. This Ranker is responsible for prioritizing agents in subsequent rounds, thereby influencing the selection of agents and shaping the evolution of the workflow. Through this iterative refinement process, DyLAN achieves adaptive and goal-oriented dynamic workflow management, allowing the system to focus on more effective agent compositions over time.

\begin{algorithm}[t]
\caption{Reward Calculation}
\label{alg:reward-calculation}
\begin{algorithmic}[1]
\STATE \textbf{Input:} current state $ s $, last node $ l $
\IF{$s$.next\_node == END}
    \STATE $ \text{path\_len} \gets |\text{executed\_nodes}| $
    \STATE $ \text{penalty} \gets \max(0, L_{\min} - \text{path\_len}) $
    \STATE $ r \gets R_{\text{success}} - \lambda_p \cdot \text{penalty} $
\ELSE
    \STATE $ c \gets \text{cost of } s.\text{next\_node} $
    \STATE $ \eta \gets \text{success rate of } s.\text{next\_node} $
    \STATE $ r \gets -\lambda_c \cdot c + \lambda_\eta \cdot \eta $
    \IF{$s.\text{next\_node} = s$}
        \STATE $ r \gets r - Penalty_n $ \COMMENT{Repeat node penalty }
    \ENDIF
\ENDIF
\RETURN $ r $
\end{algorithmic}
\end{algorithm}

\subsection{Cost Calculation}
To ensure a fair comparison, we align the number of nodes and role prompts used in our method with those of the baselines. For cost calculation, we consider both prompt and completion token usage, based on the pricing models of large language model service providers:
\begin{equation}
    Cost = p \times n_p + c \times n_c
\end{equation}
where $p$ denotes the unit price per prompt token, $c$ denotes the unit price per completion token, $n_p$ represents the number of prompt tokens used, and $n_c$ represents the number of completion tokens used.

\begin{table*}[htbp]
\centering
\setlength{\tabcolsep}{8pt} 
\begin{tabular}{l|c|c}
\hline
\textbf{Reward Component} & \textbf{Default Value} & \textbf{Scaling Factor} \\
\hline
Penalty for executing role & 5 & 2 \\
Penalty for re-executing role & 10 & 1 \\
Penalty for the number of edges & $n_{role}$ & 10 \\
Reward from role success rate &  $P_{success}(role)$ & 5 \\
Reward for completing the user task & 100 & 1 \\
\hline
\end{tabular}
\caption{Specific values and corresponding scaling factors for each component in the reward mechanism.}
\label{tab:rewards}
\end{table*}

\subsection{Multi-Model Ablation Study}
In this ablation study, we assess the generalization of our method by evaluating it on the Qwen3-Turbo model. As shown in Table \ref{tab:ablation}, our approach continues to deliver strong results, achieving superior performance over baseline methods on the HumanEval dataset. This demonstrates that the effectiveness of our method is not limited to a specific model architecture.

\begin{table}[htbp]
    \centering
    \setlength{\tabcolsep}{8pt} 
    \begin{tabular}{l|c|c}
        \hline
        \textbf{Datasets} & \textbf{Qwen2.5-Max} &  \textbf{Qwen3-Turbo}  \\
        \hline
        HuamanEval & 93.17 & 91.31 \\
        MBPP & 89.40 & 82.48 \\
        GSM8K & 96.36 & 94.40 \\
        Math & 89.10 & 84.50 \\
        \hline
    \end{tabular}
    \caption{This table compares the performance of our method on various benchmark datasets when applied to two different model backbones: Qwen2.5-Max and Qwen3-Turbo, demonstrating the consistency and generalizability of our approach across model architectures.}
    \label{tab:ablation}
\end{table}

\section{Proofs and Theoretical Analysis}
In this section, we present the design of the reward mechanism in our framework. We first elaborate on how the reward function is formulated, followed by a discussion on the rationale behind its design. Finally, we provide a formal proof of the convergence properties of our learning algorithm, ensuring the theoretical soundness of our approach.

\subsection{Reward Mechanism}
\subsubsection{Reward Calculation} As described in the main text, we design five distinct reward and penalty components to guide the learning process. In addition, each component is assigned a scaling factor to balance the relative contributions of rewards and penalties. Each individual reward term is computed according to Equation~\ref{eq2}:

\begin{equation}
    R_{c} = D_{c} \cdot S_{c},c\in C
\label{eq2}
\end{equation}

where $D_c$ denotes the default value of the reward component, $S_c$ represents its corresponding scaling factor, and $R_c$ is the resulting scaled reward or penalty. Let 
$C$ denote the space of all reward components. Then, the total reward for a given role at the current step—aggregated from all relevant components—is defined as in Equation~\ref{eq3}:

\begin{equation}
    R_{role} = \sum_{c\in C}R_{c}=\sum_{c\in C}D_{c} \cdot S_{c}
\label{eq3}
\end{equation}

\subsubsection{Rationale Behind the Reward Mechanism Design} At its core, estimating the role space to enable autonomous workflow construction is conceptually similar to playing a maze game. In both cases, the agent cannot receive immediate feedback or rewards from the environment during execution; instead, the reward is only obtained after completing an entire round of the task—equivalent to reaching the goal in a maze. Moreover, the objective in both scenarios is to reach the target (i.e., successfully complete the user task) in the shortest possible path. This analogy provides theoretical support for the feasibility of our approach.

Inspired by Q-Learning in game domains, we design our reward mechanism(Table~\ref{tab:rewards}) such that each executed node incurs a penalty, while the final reward is given only upon successful completion of the user task. However, unlike in a maze game, where all actions are treated uniformly, roles in our framework have different responsibilities and thus contribute differently to task resolution. We aim to encourage the use of roles that contribute more effectively to solving user tasks. Therefore, the penalty incurred for executing a role should vary depending on its contribution.

To achieve this, we introduce a node execution success rate-based reward, as formalized in Equation \ref{eq1}:

\begin{equation}
    P_{success}(role) = \frac{n_{success,role}}{n_{execute,role}}
\label{eq1}
\end{equation}

where $n_{success,role}$ denotes the number of times the role has participated in workflows that successfully completed the user task, and $n_{execute,role}$ represents the total number of times the role has been executed. This reward formulation allows our framework to dynamically adjust the decision-making process based on historical performance, encouraging the selection of high-contributing roles and improving overall workflow efficiency.

Our method aims to generate workflows that are both as short as possible and as effective as possible. The first objective—minimizing workflow length—is reflected through penalties on node execution and the total number of edges in the workflow. To this end, we introduce a dynamically configurable penalty for role execution, which can be adjusted by users depending on the task's complexity and desired efficiency.

The second objective—ensuring high-quality workflows—is achieved by selecting better-performing roles and identifying the most suitable next action for the current role. To support this, we incorporate a reward based on the role’s historical success rate, encouraging the system to favor roles that have contributed more effectively to previous task completions.

To balance the penalty for node execution and the reward from role success, we scale the former to fall within the range of 5–10 and the latter within 0–5. As a result, their sum is always non-positive, and its magnitude reflects the "quality" of the node: the more negative the value, the better the node performance. This design is crucial because if the sum were positive, it could lead to an undesirable situation where nodes that are executed more frequently receive higher cumulative rewards, potentially causing the system to learn and reinforce suboptimal behaviors by over-selecting certain roles.

Furthermore, since each role has a distinct responsibility, we theoretically do not expect the same role to be selected twice consecutively. Such repetition implies performing nearly identical actions, which is inefficient. Therefore, we also include an additional penalty for repeatedly selecting the same role, discouraging redundant behavior and promoting diversity in role utilization.

\subsection{Proof of Convergence}
The Q-value update formula is:

\begin{equation}
\begin{split}
    Q_{new}(s_t, a_t) = & Q(s_t, a_t) + \alpha [R_{s_t} + \gamma \max_{a}Q(s_{t+1},a)\\
            & - Q(s_t,a_t)]
\end{split}
\end{equation}

Among them, the update step size $\alpha$ is related to the time $t$, and $0<\alpha<1$ exists. We will rewrite it in the following form:

\begin{equation}
\begin{split}
Q_{new}(s_t, a_t) = & (1 - \alpha)Q(s_t, a_t) + \alpha [R_{s_t} + \\
            & \gamma \max_{a}Q(s_{t+1},a)]
\end{split}
\label{eq4}
\end{equation}

Then, assuming $\sum_{t}\alpha_{t}=\infty$ and $\sum_{t}\alpha_{t}^2<\infty$, subtract $Q^*(s_t, a_t)$ from both sides of the equation in Formula~\ref{eq4}, and let $\Delta_t(s,a) = Q^*(s_t, a_t)-Q(s_t, a_t)$, then

\begin{equation}
\begin{split}
\Delta_{t+1}(s,a)=&(1-\alpha_t(s,a))\Delta_t(s,a)+\alpha_t(s,a)[R_t \\
            &+\gamma \max_{a}Q(s_{t+1},a)- Q^*(s_t,a_t)]
\end{split}
\end{equation}

Then define:

\begin{equation}
    F_t(s,a)=R_t+\gamma \max_{b\in A}Q(s',b)- Q^*(s,a)
\end{equation}

Based on the probability distribution of environmental state transitions, the expected value can be obtained as:

\begin{equation}
\begin{split}
    \mathbb{E}[F_t(s,a)] = R_t+\gamma\sum_{y\in S}p(y|s,a)Q(y,b)-Q^*(s,a)
\end{split}
\end{equation}

Define operator $\mathcal{H}$:
\begin{equation}
    \mathcal{H} Q(s,a) = R_t+\gamma\sum_{y\in S}p(y|s,a)Q(y,b)
\end{equation}

So there is:
\begin{equation}
\begin{split}
     \mathbb{E}[F_t(s,a)] =  \mathcal{H} Q(s,a) - Q^*(s,a)
\end{split}
\end{equation}

So for the optimal action function $Q^*(s,a)$, according to the definition, it reaches the convergence point for the iterative operator $\mathcal{H}$ (otherwise it will be further iteratively optimized), which can be written as:
\begin{equation}
\begin{split}
    Q^*(s,a) =  \mathcal{H}Q^*(s,a)
\end{split}
\end{equation}

So,
\begin{equation}
\begin{split}
    \mathbb{E}[F_t(s,a)] = \mathcal{H}Q(s,a) -  \mathcal{H}Q^*(s,a)
\end{split}
\end{equation}

We can prove that $\mathcal{H}$ is a compression operator, so there is
\begin{equation}
\begin{split}
    ||\mathbb{E}[F_t(s,a)]||_\infty = ||\mathcal{H}Q(s,a) -  \mathcal{H}Q^*(s,a)||_\infty\\
    \leq\gamma||Q(s,a)-Q^*(s,a)||_\infty=\gamma||\Delta_t||_\infty
\end{split}
\end{equation}

Next, based on the definition of the random variable method, we can further derive:
\begin{equation}
\begin{split}
    var[F_t(s,a)] & =  \mathbb{E}[(F_t(s,a)-E^2[F_t(s,a)])] \\
                 & =\mathbb{E}[(R_t+\gamma \max_{b\in A}Q(s',b) -\mathcal{H}Q(s,a))^2] \\
                 & = var[R_t+\gamma \max_{b\in A}Q(s',b)]
\end{split}
\end{equation}

Because the reward function $R_t$ is bounded, there exists a constant $C$ that
\begin{equation}
\begin{split}
    var[F_t(s,a)] \leq C(1+||\Delta||_q^2)
\end{split}
\end{equation}

Therefore, according to the Theorem 1 in ~\cite{Convergence}, $\Delta_t$ can converge to 0, which means that $Q(s,a)$ can converge to $Q^*(s,a)$.

\section{Prompt Examples}
\subsection{Code Task}

\centering
\begin{notebox}[System Prompt]{Yellow}
You are a senior software engineer.\\
Please do your best to meet the user's requirements and collaborate with other agents to advance the task. \\
If you cannot complete it, hand it over to another node for further processing. \\
Don't assume; solve problems in collaboration when encountered. \\
If there is already a final answer, add 'FINAL ANSWER' before the reply. \\
Please specify the next target node in the output! \\

Already executed nodes: \{prev\_nodes\} \\
Available next nodes: \{next\_avail\_nodes\} \\

Requirement: The output should be concise, highlighting the next action.
\end{notebox}

\bigskip

\centering
\begin{notebox}[Algorithm Designer]{Blue}
You are the algorithm designer, responsible for algorithm design.\\ 
Based on the function signature and description, design the implementation using chain-of-thought reasoning. \\
Ensure the following:\\
 - Provide full pseudocode (input/output, parameter types, loop structures, edge case handling)\\
 - Clearly identify potential boundary conditions and how to handle them\\
 - If exceptions exist (e.g., division by zero, index out of bounds), explain mitigation strategies\\
 - Retain original doctests (if any)\\
 - Keep the output concise\\
 
Final format: /* next\_node: Programming Expert */ 
\end{notebox}

\bigskip

\centering
\begin{notebox}[Researcher(Traitor)]{Pink}
You are the researcher, skilled at problem-solving. \\
Upon receiving an issue from other nodes, use search tools to find solutions.\\ 
Respond using chain-of-thought reasoning. Keep your response as concise as possible.\\

Final format: /* next\_node: Programming Expert */
\end{notebox}

\bigskip

\centering
\begin{notebox}[Programming Expert]{Green}
You are the programming expert, a senior Python algorithm engineer who generates high-quality Python code based on requirements, plans, and feedback. \\

Responsibilities:\\
 - Strictly follow the original function signature (parameters, types, defaults, type hints)\\
 - You can refer to algorithm designs provided by other nodes, but they may not always be correct\\
 - Prioritize fixing errors if test feedback indicates failures\\
 - Handle all edge cases\\
Requirements:\\
 - Output must be a complete Python function definition wrapped in a code block\\
 - Do not include explanatory text, only the code block and the next node\\
 
Final format:\\
```python ...```\\
/* next\_node: Code Auditor*/
\end{notebox}

\bigskip

\centering
\begin{notebox}[Code Auditor]{Orange}
You are the code auditor, responsible for checking for fatal bugs that could cause runtime failure. \\
Focus on:\\
 - Any syntax errors?\\
 - Missing edge cases?\\
 - Do parameters match the docstring description?\\
 - Are type hints correct?\\
If issues are found, fixing it and return a complete Python function definition wrapped in a code block. \\
Otherwise, pass control to Test Enginee.\\

Final format:\\
```python ...```(if issues existed)\\
/* next\_node: Test Engineer */
\end{notebox}

\bigskip

\centering
\begin{notebox}[Test Engineer]{Cyan}
You are a test engineer. \\
The user will provide a function signature and its docstring. \\
You need to analyze the current code or solution for potential issues based on test data and feedback. \\
Provide additional test cases, edge conditions, etc., to consider during implementation. \\
If any potential errors are found, hand them back to another node for repair. \\
If everything looks good, return FINAL ANSWER as the next node.\\

Requirement: Response must be concise without unnecessary explanations and must specify a non-Test Engineer next node.\\

Final format: /* next\_node: Programming Expert */
\end{notebox}

\bigskip

\centering
\subsection{Math Task}
\begin{notebox}[System Prompt]{Yellow}
You are a professional mathematician who strives to meet the needs of users as much as possible, while collaborating with other assistants and using the provided tools to advance towards answering questions.\\
If you can't answer completely, it's okay, another assistant with different tools will continue to provide assistance where you stopped.\\ 
Do your best to advance the progress, and be sure not to assume that when you encounter problems, work with other assistants to solve them.\\
If you or any other assistant has a final answer or deliverable, add 'FINAL ANSWER' before your response so that the team knows to stop.\\
Provide the next node you want to go to in your output.\\
Please note that you need to fully understand the topic and its purpose.\\

Running Agent: \{prev\_nodes\}\\
Optional Agent: \{next\_avail\_nodes\}
\end{notebox}

\bigskip

\centering
\begin{notebox}[Math Solver]{Blue}
You are MathSolverAgent, also a math expert. \\
You will be given a math problem and hints from other agents.\\
Give your own solving process step by step based on hints. \\
If you think it's simple enough for you to solve, you use the chain of thought technique to think step by step to calculate the answer, 
and then the last line of your output contains only the final result without any units, for example: The answer is 140\\.\\
The result you provide should be as integer as possible.\\
If your task is somewhat complex, you can use the chain of thought technique to write Python programs. \\
You don't need to complete the program writing, you just need to provide the algorithm design, and then let ProgrammingExpertAgent complete the program writing.\\

At the end of the output, provide the next desired node in the following format:\\
/* next\_node: ProgrammingExpertAgent */
\end{notebox}

\bigskip

\centering
\begin{notebox}[Mathematical Analyst]{Pink}
You are MathematicalAnalystAgent, also a mathematical analyst. \\
You will be given a math problem, analysis and code from other agents. \\
You need to first analyze the problem-solving process step by step, where the variables are represented by letters. \\
Then you substitute the values into the analysis process to perform calculations and get the results.\\
The last line of your output contains only the final result without any units, for example: The answer is 140\\.\\
The result you provide should be as integer as possible.\\
At the end of the output, provide the next desired node in the following format:\\

/* next\_node: InspectorAgent */
\end{notebox}

\bigskip

\centering
\begin{notebox}[Programming Expert]{Green}
You are ProgrammingExpertAgent, also a programming expert.\\
You will be given a math problem, analysis and code from other agents. \\
Integrate step-by-step reasoning and Python code to solve math problems.\\
Analyze the question and write functions to solve the problem. \\
The function should not take any arguments and use the final result as the return value.\\
The last line of code calls the function you wrote and assigns the return value to the \(answer\) variable.\\
Use a Python code block to write your response. For example:\\```python\\def fun():\\ x = 10\\ y = 20\\ return x + y\\answer = fun()\\```\\ 
Do not include anything other than Python code blocks in your response.\\
At the end of the output, provide the next desired node in the following format:\\

/* next\_node: MathematicalAnalystAgent */
\end{notebox}

\bigskip

\centering
\begin{notebox}[Inspector]{Orange}
You are InspectorAgent, also an Inspector. \\
You will be given a math problem, analysis and code from other agents.\\
You need to choose between the following two situations based on the situation:\\
- Check whether the logic/calculation of the problem solving and analysis process is correct(if present).\\
If the code have fatal problem, give your own solving process step by step based on hints.\\
The last line of your output contains only the final result without any units, for example: The answer is 140\\.\\
The result you provide should be as integer as possible.\\
- Check whether the code corresponds to the solution analysis(if present).\\
The last line of code calls the function you wrote and assigns the return value to the \(answer\) variable.\\
Use a Python code block to write your response. For example:\\ ```python\\ def fun():\\ x = 10\\ y = 20\\ return x + y\\ answer = fun()\\```\\
If the answer is correct, you do not need to provide a Python implementation.\\

At the end of the output, provide the next desired node in the following format:\\
/* next\_node: END */
\end{notebox}

\bigskip

\subsection{General Task}
\begin{notebox}[System Prompt]{Yellow}
You are an expert. Please try your best to meet the needs of users and cooperate with other assistants to promote the task.\\
If you can't answer completely, it's okay, another assistant with different tools will continue to provide assistance where you stopped. \\
Do your best to advance the progress, and be sure not to assume that when you encounter problems, work with other assistants to solve them.\\
If you or any other assistant has a final answer or deliverable, add 'FINAL ANSWER' before your response so that the team knows to stop.\\
Provide the next node you want to go to in your output!!!\\
**Select appropriate nodes according to different user task types.Please note that you need to fully understand the topic and its purpose.**\\
Running node: {prev\_nodes}\\
Optional nodes: {next\_avail\_nodes}
\end{notebox}

\bigskip

\begin{notebox}[Planner Prompt]{Blue}
You are the leader of the team. Your team is responsible for solving mathematical or code tasks. \\
If the task given to you by the user is a code task, based on the task, design the implementation using chain-of-thought reasoning.\\
Ensure the following:\\
 - Provide full pseudocode (input/output, parameter types, loop structures, edge case handling)\\
 - Clearly identify potential boundary conditions and how to handle them\\
 - If exceptions exist (e.g., division by zero, index out of bounds), explain mitigation strategies\\
 - Keep the output concise\\
 - Don't goto END or FINAL\_ANSWER directly.\\
 
If the task given to you by the user is a mathematical task, based on the task, decide whether to solve it with mathematical techniques or design an algorithm approach.\\
If it's a mathematical problem, give your own solving process step by step. \\
If it's an algorithmic problem, design the approach and provide pseudocode.\\
At the end of your response, provide the final result (for math) or approach outline (for algorithms), and specify the next node.\\
The last line of your output contains only the final result without any units for math problems, for example: The answer is 140.\\
For algorithmic problems, provide a brief outline of the approach.\\
**You don't need programming**\\
 At the end of the output, provide the next desired node in the following format:\\
/* next\_node: Programming Expert */
\end{notebox}

\bigskip

\begin{notebox}[Programming Expert]{Green}
You are ProgrammingAgent, a senior Python algorithm engineer who generates high-quality Python code based on requirements, plans, and feedback.\\
You will be given a task, analysis, and code from other agents.\\

Responsibilities:\\
 - Strictly follow the original function signature (parameters, types, defaults, type hints)\\
 - You can refer to algorithm designs provided by other nodes, but they may not always be correct\\
 - Prioritize fixing errors if test feedback indicates failures\\
 - Handle all edge cases\\
Requirements:\\
 - Output must be a complete Python function definition wrapped in a code block\\
 - Do not include explanatory text, only the code block and the next node\\

If the task is a math task, ensure the following:\\
    -The last line of code calls the function you wrote and assigns the return value to the \(answer\) variable. For example:\\ '''python\\def fun():\\ x = 10\\ y = 20\\ return x + y\\answer = fun()\\'''\\
If the task is a code task, ensure the following:\\
    -Try not to output anything except code blocks.\\
    -Don't have 'answer=fun()' in your code blocks\\
Do not include anything other than Python code blocks in your response.\\

At the end of the output, provide the next desired node in the following format:\\
/* next\_node: Mathematical Analyst */
\end{notebox}

\bigskip

\begin{notebox}[Mathematical Analyst]{Pink}
You are AnalystAgent, a problem analyst who can handle both mathematical analysis and code analysis.\\
You will be given a task, previous analysis, code, and results from other agents.\\
If the task involves mathematical analysis, analyze the problem-solving process step by step, where the variables are represented by letters.\\
Then substitute the values into the analysis process to perform calculations and get the results.\\
If the task involves code analysis, analyze the code logic, identify potential issues, and suggest improvements.\\
The last line of your output contains only the final result without any units for math problems, for example: The answer is 140.\\
At the end of the output, provide the next desired node in the following format:\\
/* next\_node: Inspector */
\end{notebox}

\bigskip

\begin{notebox}[Inspector]{Orange}
You are InspectorAgent, an inspector who can check both mathematical reasoning and code correctness.\\
You will be given a task, analysis, code, and results from other agents.\\
Check whether the logic/calculation of the problem solving and analysis process is correct (if present).\\
Check whether the code corresponds to the solution analysis (if present).\\
If there are issues, provide corrections and explain the problems.\\
If the answer is correct and the code is working properly, you can finish the task.\\
At the end of the output, provide the next desired node in the following format:\\
/* next\_node: END */
\end{notebox}

\bigskip

\begin{notebox}[Code Auditor]{Violet}
You are CodeAuditorAgent, a code audit expert.\\
You will be given code that needs to be checked for bugs, vulnerabilities, and optimization opportunities.\\
Analyze the code and identify any issues.\\
If issues are found, provide corrected code.\\
If the code is correct, confirm its correctness.\\
At the end of the output, provide the next desired node in the following format:\\
/* next\_node: END */
\end{notebox}

\bigskip

\begin{notebox}[Test Engineer]{Cyan}
You are the TestEngineerAgent, a test analyst. \\
The user will provide a function signature and its docstring. \\
You need to analyze the current code or solution for potential issues based on test data and feedback. \\
Provide additional test cases, edge conditions, etc., to consider during implementation. \\
If any potential errors are found, hand them back to another node for repair. \\
If everything looks good, return FINAL ANSWER as the next node.\\

Requirement: Response must be concise without unnecessary explanations and must specify a non-test-engineer next node.\\

Final format: /* next\_node: Code Auditor */
\end{notebox}

\newpage

\bibliography{aaai2026}

\end{document}